\documentclass{article}

\usepackage{PRIMEarxiv}

\usepackage[utf8]{inputenc} % allow utf-8 input
\usepackage[T1]{fontenc}    % use 8-bit T1 fonts
\usepackage{hyperref}       % hyperlinks
\usepackage{url}            % simple URL typesetting
\usepackage{booktabs}       % professional-quality tables
\usepackage{amsfonts}       % blackboard math symbols
\usepackage{nicefrac}       % compact symbols for 1/2, etc.
\usepackage{microtype}      % microtypography
\usepackage{lipsum}
\usepackage{fancyhdr}       % header
\usepackage{graphicx}       % graphics
\graphicspath{{media/}}     % organize your images and other figures under media/ folder

\usepackage{amsmath,amsfonts}
\usepackage{algorithmic}
\usepackage{graphicx}
\usepackage{graphics}
\usepackage{multirow}
\usepackage[ruled,vlined,linesnumbered,algo2e,noend]{algorithm2e}
\usepackage{algorithm}
\usepackage{array}
\usepackage[caption=false,font=normalsize,labelfont=sf,textfont=sf]{subfig}
\usepackage{textcomp}
\usepackage{stfloats}
\usepackage{url}
\usepackage{verbatim}
\usepackage{cite}
\usepackage[ruled,vlined,linesnumbered,algo2e,noend]{algorithm2e}
\usepackage{adjustbox}

\DeclareMathOperator{\transformer}{Transformer\_Encoder}
\DeclareMathOperator{\embedding}{Embedding}

\title{SubgroupTE: Advancing Treatment Effect Estimation with Subgroup Identification}

\author{
  Seungyeon Lee, Ruoqi Liu \\
  The Ohio State University \\
  \texttt{\{lee.10029,liu.7324\}@osu.edu} \\
   \And
  Wenyu Song \\
  Harvard Medical School \\
  \texttt{wsong@bwh.harvard.edu} \\
  \And
  Lang Li \\
  The Ohio State University \\
  \texttt{Lang.Li@osumc.edu} \\
  \And
  Ping Zhang\thanks{Corresponding author} \\
  The Ohio State University \\
  \texttt{zhang.10631@osu.edu}
}

\begin{document}
\maketitle

\begin{abstract}
Precise estimation of treatment effects is crucial for evaluating intervention effectiveness. While deep learning models have exhibited promising performance in learning counterfactual representations for treatment effect estimation (TEE), a major limitation in most of these models is that they treat the entire population as a homogeneous group, overlooking the diversity of treatment effects across potential subgroups that have varying treatment effects. This limitation restricts the ability to precisely estimate treatment effects and provide subgroup-specific treatment recommendations. In this paper, we propose a novel treatment effect estimation model, named SubgroupTE, which incorporates subgroup identification in TEE. SubgroupTE identifies heterogeneous subgroups with different treatment responses and more precisely estimates treatment effects by considering subgroup-specific causal effects. In addition, SubgroupTE iteratively optimizes subgrouping and treatment effect estimation networks to enhance both estimation and subgroup identification. Comprehensive experiments on the synthetic and semi-synthetic datasets exhibit the outstanding performance of SubgroupTE compared with the state-of-the-art models on treatment effect estimation. Additionally, a real-world study demonstrates the capabilities of SubgroupTE in enhancing personalized treatment recommendations for patients with opioid use disorder (OUD) by advancing treatment effect estimation with subgroup identification.
\end{abstract}

\keywords{deep learning, treatment effect estimation, subgroup analysis, opioid use disorder}

\section{Introduction}
The primary measure for assessing the effectiveness of interventions is based on the treatment effect (TE), which quantifies the difference in outcomes between the treatment and control groups. The precise estimation of treatment effects is, therefore, of great importance in accurately evaluating the intervention. In recent years, many deep learning models have been proposed and demonstrated remarkable contributions in treatment effect estimation (TEE) \cite{curth2021inductive,schwab2020learning, shalit2017estimating,shi2019adapting,nie2021vcnet,zhang2022exploring}. These models leverage the power of neural networks to capture intricate relationships among covariates, treatments, and outcomes. Nevertheless, most of these models overlook the existence of heterogeneous subgroups with distinct treatment effects and characteristics,  leading to limitations in providing more accurate subgroup-specific estimations and treatment recommendations.

Recognizing the significance of heterogeneity in treatment effects across subgroups is critical, especially considering the diverse nature of populations in real-world situations. Consequently, it is essential to understand and analyze the variability of treatment effects to advance precision treatment. Subgroup analysis aims to identify heterogeneous subgroups with various treatment responses to account for this variability. Nevertheless, existing subgroup identification methods  \cite{yang2022tree,loh2016identification, foster2011subgroup, lee2020causal, lee2020robust, argaw2022identifying, nagpal2020interpretable} have challenges to be addressed. Firstly, most of these methods utilize conventional machine learning models that may encounter challenges when attempting to learn from high-dimensional data with intricate and non-linear relationships among covariates, treatments, and outcomes. Secondly, they highly depend on a one-time pre-estimation of treatment effects to find subgroups whose individuals have similar responses, which can lead to the sub-optimal problem if the pre-estimated effects are inaccurate.

To address the issues, we propose a treatment effect estimation model, called SubgroupTE, that incorporates subgroup identification to find heterogeneous subgroups and learns subgroup-specific causal effects. SubgroupTE advances the estimation of treatment effects by the identified subgroups rather than estimating effects for the entire population. Furthermore, we design an expectation–maximization (EM)-based training process that iteratively optimizes estimation and subgrouping networks to improve estimation and subgroup identification simultaneously. The EM-based training process effectively overcomes the limitations of the pre-estimation step in subgroup analysis, enhancing the overall performance of the model. The SubgroupTE comprises three primary networks: A feature representation network that extracts useful representations from the input data, a subgrouping model that identifies heterogeneous subgroups, and a subgroup-informed prediction network that estimates the treatment effects by considering subgroup-specific causal effects. Specifically, the feature representation network consists of an embedding layer and a Transformer encoder to learn non-linear representations. The subgrouping model pre-estimates the treatment effects and finds subgroups based on the estimated effects. The subgroup-informed prediction network estimates the treatment effects using latent features from the feature representation network and subgroup probabilities from the subgrouping model to consider subgroup-specific causal effects.

Our contributions can be outlined as follows:
\begin{itemize}
    \item We propose a novel treatment effect estimation model, SubgroupTE, that incorporates subgrouping to identify heterogeneous subgroups and more precisely estimates treatment effects by learning subgroup-specific causal effects.
    \item We introduce an EM-based training process to iteratively optimize prediction and subgrouping, enhancing both estimation and subgroup identification.
    \item We demonstrate that considering the heterogeneity of responses within the population leads to more precise predictions compared to estimating treatment effects for the entire population.
    \item We demonstrate the potential of our method in enhancing treatment recommendations in real-world scenarios.    
    \item We show that SubgroupTE helps to identify the variables contributing to the improvement of treatment effects.    
\end{itemize}

Note that SubgroupTE was introduced in our previous conference paper. The main differences between this paper and the previous conference paper are summarized as follows:
\begin{itemize}
 \item We perform a robust quantitative analysis, leveraging a semi-synthetic dataset, with state-of-the-art treatment effect and subgrouping models to demonstrate the superiority of SubgroupTE in both treatment effect estimation and subgrouping tasks.
\item We conduct an ablation study to evaluate our EM-based training process, demonstrating the strength of iterative optimization for subgrouping and treatment estimation.
\item We conduct a sensitivity analysis to assess the impact of the number of subgroups on the prediction performance, demonstrating the significance of identifying subgroups with similar treatment responses and integrating them into the estimation process. 
\item Code and data are available in a code repository\footnote{ \url{https://github.com/yeon-lab/SubgroupTE}}.
\end{itemize}

\section{Related works}\label{sec2}
This section briefly reviews existing work relevant to our study, including treatment effect estimation and subgroup analysis for treatment effects.

\subsection{Treatment effect estimation.}
There have been numerous efforts to leverage the power of neural networks in learning counterfactual representations for treatment effect estimation. To successfully apply neural networks for causal inference, it is crucial to design the network structure thoughtfully. Specifically, neural network models are designed to differentiate the treatment variable from other covariates to preserve the treatment-related information in the latent representation. To prevent the loss of treatment information, some previous works \cite{shi2019adapting,curth2021inductive,schwab2020learning, shalit2017estimating} employ a strategy where covariates from different treatment groups are assigned to separate branches. For example, DragonNet~\cite{shi2019adapting} utilizes a shared feature network along with three distinct auxiliary networks, which predict the propensity score, and treated and control outcomes, respectively. By segregating the covariates based on the treatment and utilizing dedicated auxiliary networks, it effectively captures and accounts for the impacts of different treatments. In another effort to avoid loss of treatment information, VCNet~\cite{nie2021vcnet} maps a real-value treatment to the $n$-dimensional vector with a mapping function to preserve the treatment information. TransTEE~\cite{zhang2022exploring} leverages Transformer architecture to capture the interaction between the input treatment and covariates.

Deep learning models have exhibited successful performance in TEE. However, a major limitation in most of these models is that they treat the entire population as a single group, overlooking the diversity of treatment effects across potential subgroups that have varying treatment effects. This can hinder the capacity to precisely estimate treatment effects and provide personalized treatment recommendations for specific subgroups within the population. On the contrary, our novel approach tackles the challenges by incorporating subgroup identification and treatment effect estimation, thereby advancing estimation by considering subgroup-specific causal effects.

\subsection{Subgroup analysis} 
Subgroup analysis for causal inference aims to find subgroups whose individuals have similar treatment responses and/or characteristics. These subgroup analyses are categorized into three distinct groups \cite{wang2022causal}.

\textbf{Subgrouping with pre-defined hypotheses} The approach focuses on analyzing the treatment effect within specific subgroups that are identified based on a priori hypotheses. These hypotheses are formed by considering factors or characteristics that are believed to influence the treatment response. 

\textbf{Directly subgrouping without hypotheses and estimation} The approach directly finds subgroups based on statistics, latent patterns, and/or distributions in the data without a priori hypotheses and estimating treatment effect.

\begin{figure*}[t]
\centering
\includegraphics[width=0.9\linewidth]{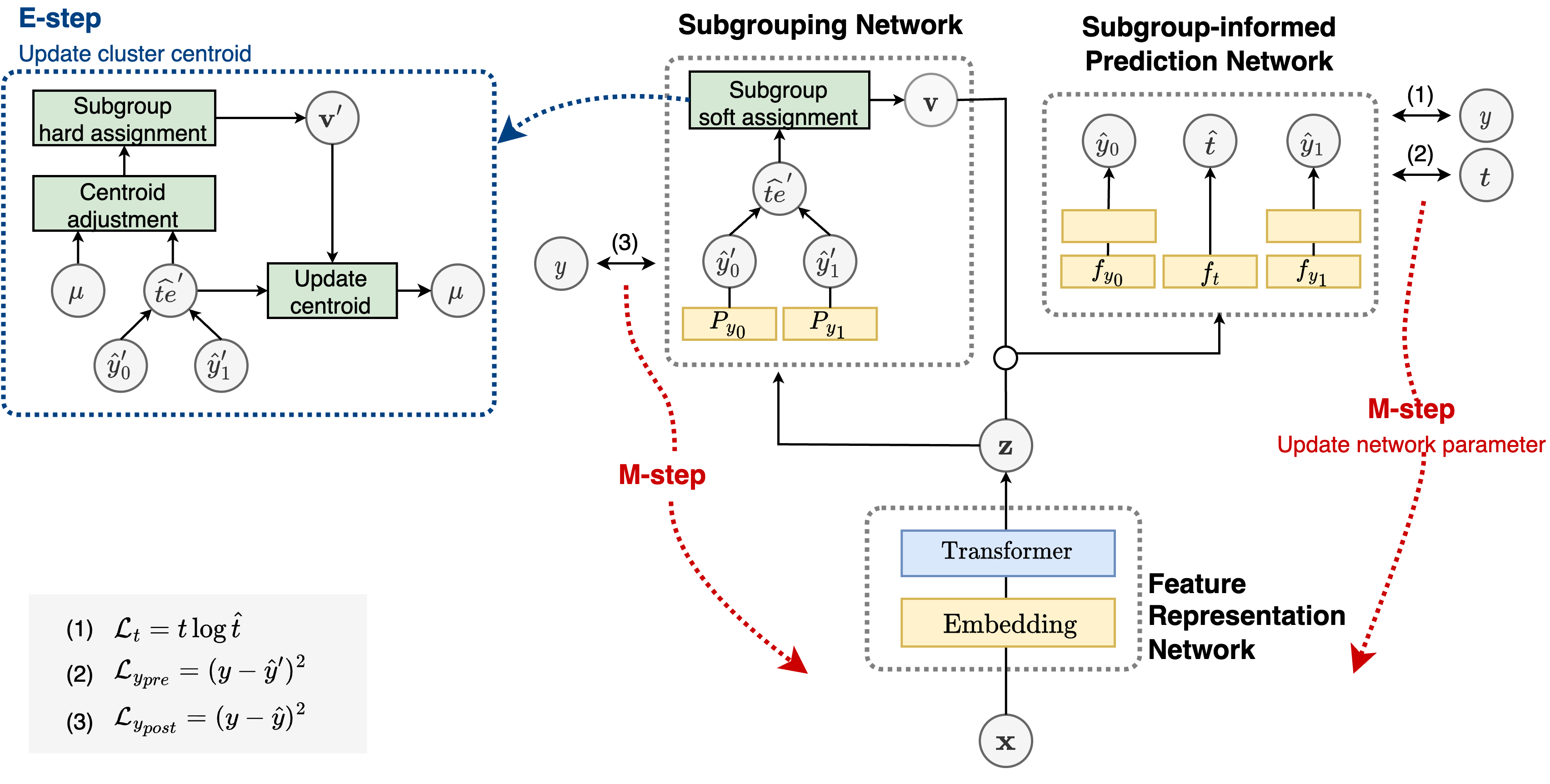}
\caption{An overview of SubgroupTE.}
\label{fig1}
\end{figure*}

\textbf{Subgrouping with estimated potential outcomes} The approach first pre-estimates the potential outcomes and then finds subgroups by maximizing treatment effect heterogeneity/homogeneity across/within subgroups. For example, \cite{yang2022tree,loh2016identification, foster2011subgroup, lee2020causal, lee2020robust, argaw2022identifying} recursively split covariate space into subgroups and find optimal subgroups by maximizing treatment effect heterogeneity. HEMM\cite{nagpal2020interpretable} utilizes Gaussian mixture distributions of subgroups to generate the input, resulting in subject-specific subgroup probabilities. HEMM employs a pre-trained estimation model to learn the parameters of the distributions. The effectiveness of this approach is strongly dependent on the accuracy of the one-time pre-estimation, which may pose challenges in some situations.

Our proposed model belongs to the third category of subgrouping approaches, but it tackles the challenges associated with the one-time pre-estimation. To overcome the challenges, we employ an EM-based training process that iteratively optimizes both subgrouping and treatment effect estimation networks until they converge. This iterative approach leads to enhanced subgroup identification and more precise estimation for each subgroup.
% 1. iteratively subgroup and TE estimation
% 2. iterative optimization is better than exisitng works

\section{Methodology}\label{sec3}
In this section, we introduce the basic notations, problem formulation, and the proposed framework. We use upper-case and bold letters (e.g., $\textbf{X}$) for matrices, lower-case and bold letters (e.g., $\textbf{x}$) for vectors, and lower-case letters (e.g., $x$) for a scalar. Table~\ref{tb1} summarizes the notations used in our study.

\subsection{Problem statement}

\noindent\textbf{Treatment effect estimation}
We consider a setting in which we are given a dataset $D \equiv \{(\mathbf{x}_i,t_i,y_i)\}^N_{i=1}$, where $N$ is the number of observed samples. $\mathbf{x}_i \in \mathbb{R}^p$ and $t_i\in\mathbb{R}$ represent $p$ pre-treatment covariates and treatment variable, respectively. $t_i$ is binary when binary treatment setting. The potential outcome $y_i$ indicates the treatment response for $t_i$ of the $i$-th sample. The propensity score is defined as the conditional probability of the treatment assignment given the observed covariates, represented as $p(t=1|\mathbf{x})$. Under the potential outcomes framework\cite{rubin1974estimating}, the treatment effect is defined as the difference between the treated and control outcomes, represented as $\mathbb{E}[Y(1) - Y(0) |\mathbf{x}]$ at a given value of $\mathbf{x}$. The goal of the prediction model for causal inference is to accurately estimate the treated and control outcomes. With the training data, which includes factual samples $\mathbf{x}_i$, $y_i$, and $t_i$, the model is trained to estimate the factual outcome $y_i$, given $\mathbf{x}_i$ and $t_i$.

Following \cite{lechner2001identification}, we depend on the assumption of unconfoundedness, which are as follows: (1) Conditional Independence Assumption: The assignment of treatment is independent of the outcome, given the pre-treatment covariates. (2) Common Support Assumption: There is a nonzero probability of the treatment assignment for all samples. (3) Stable Unit Treatment Value Assumption: The observed outcome of each unit remains unaffected by the assignment of treatments to other units. These assumptions are essential in treatment effect estimation as they provide the necessary conditions for unbiased and consistent estimation of causal effects. The assumptions form the basis for our methodology. 

\noindent\textbf{Assumption 1} (Ignorability/Unconfoundedness) implies that there are no hidden confounders, such that $Y(0), Y(1) \perp  T|X$.

\noindent\textbf{Assumption 2}
(Positivity/Overlap) implies that the treatment assignment is non-deterministic, such that $0<\pi(t|\mathbf{x})<1$.

\begin{table}[t]
\caption{Notation definition}
\centering
\adjustbox{max width=\linewidth}{
\begin{tabular}{ll}
\toprule
Notation & Description \\ \hline
$D \equiv \{ \left(\textbf{x}_{i}, y_i, t_i\right) \}_{i=1}^{|D|}$ & input dataset \\ 
$\textbf{x}_i$ & covariate of the $i$-th patient \\
$y_{i}$ & factual outcome\\
$t_{i}$ & treatment assignment\\ 
$\hat{y}_{i}$ & prediction of factual outcome\\
$\hat{t}_{i}$ & prediction of treatment assignment\\ 
$\hat{te}'_{i}$ & prediction of pre-subgrouping treatment effect \\ 
$\hat{y}'_0,\hat{y}'_1$ & prediction of pre-subgrouping control and treated outcomes \\
$\hat{y}_0,\hat{y}_1$ & prediction of post-subgrouping control and treated outcomes\\
$\textbf{z}_i$ & latent representations from $Q_{\phi}$\\ 
$\textbf{v}_i$ & subgroup probability vector\\ 
$K$ & the number of subgroups \\ 
$\mu_k$ & $k$-th cluster centroid \\ 
$Q_{\phi}$ & feature representation network \\ 
$P_{y_0}$ & subgrouping network for control outcome estimation\\ 
$P_{y_1}$ & subgrouping network for treated outcome estimation\\ 
$f $ & subgroup-informed prediction network\\ 
$\alpha, \beta, \gamma $ & weights to control losses \\
\hline
\end{tabular}}
\label{tb1}
\end{table}

\subsection{Proposed model}
The proposed model, SubgroupTE, comprises three primary networks: the feature representation network, the subgrouping network, and the subgroup-informed prediction network. The feature representation network, based on the encoder network of the Transformer model, extracts useful representations from the input data. The subgrouping network pre-estimates the treatment effects and subgroups the data based on the estimated outcomes, generating a subgroup probability vector representing the likelihood of belonging to subgroups. The subgroup-informed prediction network performs the final estimation of the treatment effects, leveraging the latent representation from the feature representation network and subgroup probabilities as input. The SubgroupTE is optimized by the EM-based training, which iteratively updates the cluster centroids for subgrouping and network parameters. 

Fig. \ref{fig1} provides a comprehensive overview of the SubgroupTE framework, highlighting the flow of information and the interactions between the different components. The training process of the proposed method is provided in Algorithm \ref{alg:fatdm}.

\subsubsection{\textbf{Feature representation network}} To extract latent representations from the input, we employ an embedding layer and encoder network of the Transformer model to construct a feature representation network $Q_{\phi}$. The encoder network consists of self-attention and feed-forward neural networks. Self-attention enables a model to focus on different parts of the input during the encoding process. It allows the model to selectively attend to relevant information at each position

Given the $i$-th sample $\textbf{x}_i$, it is first fed into the embedding layer, and the resulting output is then input into $\transformer$ to extract the latent features of $\textbf{x}_i$.

\begin{align}
    \textbf{z}_i &= Q_{\phi}(\textbf{x}_i) \\
    & = \transformer\left(\embedding(\textbf{x}_i)\right) \nonumber
\end{align}\label{eq1}

\begin{algorithm2e}[t]
\caption{Training process for SubgroupTE}
\label{alg:fatdm}
\SetAlgoLined
\KwIn{Dataset $\mathcal{D}$, The number of subgroups $K$}
\KwOut{Feature representation network $Q_{\phi}$, Subgrouping networks $P_{y_0},P_{y_1}$, Subgroup-informed prediction network $f$,
Estimated treatment effect, Subgroups}

\setcounter{AlgoLine}{0}
\SetKwFunction{size}{count\_atoms}
\SetKwFunction{bond}{get\_brics\_bonds}
\SetKwFunction{brk}{break\_molecule}
\SetKwFunction{len}{length}
\SetKw{and}{and}
\SetKw{or}{or}
\SetKw{not}{not}
\SetKw{true}{true}
\SetKw{false}{false}
\SetKwProg{myproc}{Procedure}{}{}

\For{epoch $=1$ to E}{
\For{batch $\textbf{b} = \{ \left(\mathbf{x}_{i}, y_i, t_i\right) \}_{i=1}^{|\textbf{b}|}$ in $\mathcal{D}$}{
\textbf{\textit{E-step;}} \\
$\mathbf{z} = Q_\theta(\mathbf{x})$ \\
$\widehat{te}' = P_{1}(\textbf{z}) - P_{0}(\textbf{z})$ \\
Adjust cluster centroids $\mu$ using $\widehat{te}'$ by Eq. (\ref{eq4}) \\
Assign clusters $\mathbf{v}'$ by Eq. (\ref{eq5})\\
Update $\mu$ by Eq. (\ref{eq6})\\
\vspace{1em}
\textbf{\textit{M-step;}} \\
Compute subgroup probability $\mathbf{v}$ by Eq. (\ref{eq3})\\
$\hat{y},\hat{t}=f(\mathbf{v},\mathbf{z})$ \\
Update networks using $\hat{y},\hat{t}$ by Eq. (\ref{eq7})
}\textbf{end}
}\textbf{end}
\end{algorithm2e}

\subsubsection{\textbf{Subgrouping model}} The goal of the subgrouping model is to identify subgroups of patients that have enhanced or diminished treatment effects. Considering the diverse treatment responses observed among individuals, the subgrouping model aims to capture and account for this heterogeneity by identifying distinct subgroups. To achieve this, the subgrouping network first estimates the control and treated outcomes, denoted as $\hat{y}'_{0}$ and $\hat{y}'_{1}$ (pre-subgrouping estimations), respectively. This estimation is performed using two distinct one-layer feedforward networks: $P_{y_0}$ and $P_{y_1}$. These networks take the covariates $\textbf{z}_i$ as input and predict potential outcomes for the control and treated samples, respectively. The networks have architectures that map the covariates from $\mathbb{R}^{N \times |\mathbf{z}|}$ to $\mathbb{R}^{N \times 1}$.

The pre-subgrouping treatment effect is then computed for each data sample. Based on the homogeneity of the estimated treatment effects, the subgrouping model assigns a subgroup probability vector to each data sample. The subgroup probability vector indicates the likelihood of a data sample belonging to each subgroup. It reflects the similarity in treatment effects within subgroups.

The subgroup probability vector, expressed as $\mathbf{v}\in\mathbb{R}^K$, with $K$ representing the number of subgroups, is calculated by measuring the distance between the subgroup centroids and pre-subgrouping treatment effect. Given the $i$-th sample, the distance $d_i^k$ is computed as the Euclidean distance between the centroid $\mu_k$ of subgroup $k$ and $\widehat{te}_i$ as follows:

%\liucomment{missing equation no.}
\begin{equation*}
    \widehat{te}'_i = P_{1}(\textbf{z}_i) - P_{0}(\textbf{z}_i)
\end{equation*}

\begin{equation}\label{eq2}
    d_i^k= \left \|\widehat{te}'_i-\mu_k \right \|_2
\end{equation}
where $\left \| \cdot \right \|_2$ denotes the Euclidean norm.

The probability $v_i^k$ that the $i$-th sample belongs to the subgroup $k$ is then calculated using the distance $d_i^k$:

\begin{equation}\label{eq3}
    v_{i,k}=\frac{\exp^{-d_i^k}}{\sum_{j=1}^K \exp^{-d_i^j}}
\end{equation}

The probability $v_{i,k}$ is obtained by exponentiating the negative distance $d_i^k$ and normalizing it across all subgroups. The closer the treatment effect $\widehat{te}'_i$ is to the centroid $\mu_k$, the higher the probability $v_{i,k}$ for that subgroup. The subgroup probability vector $\mathbf{v}_i$ is used along with the latent features from the feature representation network as input to the subgroup-informed prediction network. With these subgroup probabilities, the subgroup-informed prediction network can leverage the subgroup information for the treatment effect estimation process, allowing for learning subgroup-specific causal effects.

\subsubsection{\textbf{Subgroup-informed prediction network}}
The SubgroupTE framework aims to identify heterogeneous subgroups of patients with different treatment effects and improve the precision of outcome estimation by considering subgroup-specific causal effects. To achieve this, the representation vector $\textbf{z}_{i}$ and the subgroup probability vector $\textbf{v}_{i}$ are concatenated and used as input for the subgroup-informed prediction network.

To ensure the preservation of treatment information in the high-dimensional latent representation, we adapt a strategy where covariates from different treatment groups are assigned to separate branches. The subgroup-informed prediction network $f$ comprises three separate feedforward networks: $f_{y_0}$, $f_{y_1}$, and $f_{t}$. These networks are responsible for predicting control and treated outcomes, and treatment assignment, respectively. By employing this approach, we ensure that the treatment variable is effectively distinguished from other covariates, thus preserving and incorporating the treatment information into our model. The outputs of the subgroup-informed prediction network are expressed as follows:

\begin{align}
    \hat{y}_{0,i}&= f_{y_0}(\textbf{v}_i, \textbf{z}_i) \\
    \hat{y}_{1,i}&= f_{y_1}(\textbf{v}_i, \textbf{z}_i) \\
    \hat{t}&= f_{t}(\textbf{v}_i, \textbf{z}_i) 
\end{align}
where $\hat{y}_{0,i}$ and $\hat{y}_{1,i}$ represent the post-subgrouping estimation of control and treated outcomes for $i$-th sample, respectively, and $\hat{t}_i$ represents the predicted treatment assignment. 
By incorporating the subgroup probability vector $\textbf{v}_{i}$ along with the representation vector $\textbf{z}_{i}$ as inputs to the subgroup-informed prediction network, the SubgroupTE framework allows for subgroup-specific treatment effect estimation. This enables a better understanding of the heterogeneity in treatment effects across different subgroups.

\subsubsection{\textbf{Optimization and initialization}}

To optimize the proposed SubgroupTE model, we employ an expectation–maximization (EM)-based training process, which iteratively optimizes the network parameters and cluster centroids. 

In the \textbf{E-step}, the cluster centroids are updated while all network parameters are frozen. The K-means algorithm is utilized to assign patients into $K$ subgroups based on the homogeneity of the estimated treatment effects and update cluster centroids. The loss function is as follows:

\begin{equation}
    \min_{M\in\mathbb{R}^{K \times 2}} \sum_{i}\left \| \widehat{te}'_i- v'_i\times M \right \|_2, v'_i \in \mathbb{R}^{K}
\end{equation}
where $v'_i$ is a hard assignment vector that assigns each data sample $i$ to one of the $K$ clusters. The centroid matrix $M$ represents the centroids of each cluster, with the $k$-th row, denoted as $\mu_k$. 

Data points are assigned to the nearest clusters based on the distance to their centroids. However, as network parameters are trained, the distribution of the feature space in the network, that is the pre-subgrouping treatment effect estimation, may shift, resulting in discrepancies with the previously updated centroid distribution. This causes the distance between the existing centroids and the data points mapped in the updated feature space to increase. Consequently, some clusters may no longer contain any data points, leading to a reduction in the number of clusters. 

To address this issue, we first shift the distribution of the existing centroids to the new feature space before assigning data points to clusters. We accomplish this by computing the Kernel Density Estimation (KDE) of the distribution between the existing centroids and the data points mapped in the new feature space and then updating the centroids using KDE. Specifically, the proposed method includes the following steps. First, we obtain the hidden features in the updated network given the data points. Next, for each centroid, we calculate the KDE between the distributions of the centroid and hidden features. The KDE represents the shift in distribution caused by the changes in the network parameters. Based on the KDE, we update each centroid's position by adjusting it according to the difference in distribution. This adjustment allows the centroids to adapt to the new feature space and ensures that they are representative of the underlying feature distribution. Finally, the data points are assigned to the updated centroids.

The centroids are adjusted as follows:
\begin{equation*}
\textrm{Kernel}(\widehat{te}'_i,\mu_k) = \frac{e^{-\frac{1}{2} ((\widehat{te}'_i-\mu_{k})/\cdot h)^2}}{\sum_i e^{-\frac{1}{2} ((\widehat{te}'_i-\mu_{k})/\cdot h)^2}}
\end{equation*}

\begin{equation*}
\textrm{Diff}(\widehat{te}', \mu_k) = \sum_i \\ 
\textrm{Kernel}(\widehat{te}'_i,\mu_k)\cdot (\widehat{te}'_i-\mu_k)
\end{equation*}

\begin{equation}\label{eq4}
\mu^*_k = \mu_k + \textrm{Diff}(\widehat{te}', \mu_k)
\end{equation}

The kernel represents a weight indicating how close each data point is to the cluster center, where a data point closer to the cluster center has a larger value, while a data point farther away has a smaller value. The difference between the data points and each cluster center is then multiplied by its weight, and the sum of these weighted data points over all data points yields the density value. Therefore, $\textrm{Diff}(\cdot)$ indicates the density distribution of the data points with respect to cluster centers. $\mu_k + \textrm{Diff}(\cdot)$ adjusts the current position of the cluster center to move towards high-density regions in the feature space.

By aligning the centroids' distribution with the new feature space through KDE-based adjustments before assigning data points to clusters, the proposed method mitigates the issue of decreasing cluster numbers caused by changes in the feature space distribution. The centroid update procedure accounts for the shift in distribution and allows the clusters to adapt to the evolving data representation.

The centroid matrix is then updated using the following equations given a mini-batch. The hard assignment vector $v'_i$ is computed as:

\begin{equation}\label{eq5}
    v'_{i,j}= \left\{\begin{matrix}
    1, & j = \textrm{argmin}_{k=\{1,...,K\}} {\left \| \widehat{te}'_i-\mu^*_k \right \|_2}, \\ 
    0, & otherwise.
    \end{matrix}\right.
\end{equation}

The centroid is updated as follows:

\begin{align}\label{eq6}
\mu_{k}= \left\{\begin{matrix}
\frac{1}{|B_k|}\sum_{i\in B_k}\widehat{te}'_i,&|B_k|>0 \\ 
\mu^*_k, & otherwise.
\end{matrix}\right.
\end{align}
where $B_k$ represents all the samples assigned to $k$-th cluster such that $B_k = \{i \mid \forall i, v_{i,k}=1\}$.

In the \textbf{M-step}, the network parameters are updated. During this step, the cluster centroids are fixed, and a subgroup probability vector $\mathbf{v}$ is assigned to each sample using Eq. (\ref{eq3}). The network parameters are subsequently updated using the predictions from the Subgrouping and Subgroup-informed prediction networks. The loss function is defined as follows:

\begin{align}
\label{eq7}    \mathcal{L} & = \alpha \cdot \mathcal{L}_{t}+\beta \cdot \mathcal{L}_{y_{pre}}+\gamma \cdot \mathcal{L}_{y_{post}} \\ 
 & = \alpha \cdot \sum_{i=1}^{|D|}t_i\log{\hat{t}_i} +\beta \cdot \sum_{i=1}^{|D|}(y_i-\hat{y}'_i)^{2} + \gamma \cdot \sum_{i=1}^{|D|}(y_i-\hat{y}_i)^{2} \nonumber
\end{align}
where $\hat{y}'_{i}$ and $\hat{y}_{i}$ indicate the pre- and post-subgrouping estimation of $i$-th sample's factual outcomes from the subgrouping and subgroup-informed prediction networks, respectively. $\alpha$, $\beta$, and $\gamma$ are hyper-parameters that control the importance of the treatment assignment, and the pre- and post-subgrouping estimation losses, respectively.

\noindent\textbf{Initialization} We leverage k-means++ algorithm \cite{arthur2007k} to initialize the cluster centroid. It first assigns the first centroid to randomly selected data points from a mini-batch of the input data set. The subsequent centroid is then selected from the remaining data points, with their probabilities determined by the squared distance to the nearest existing centroids. This approach aims to push the centroids as far apart as possible while effectively covering a larger portion of the data space.

\section{Experiments}\label{sec4}
In this section, we assess the performance of the proposed SubgroupTE on two synthetic datasets and compare the results with existing models. Additionally, we demonstrate the practical effectiveness of SubgroupTE in subgroup identification using a real-world dataset.

\subsection{Datasets}
We conduct our experiments on three datasets including synthetic, semi-synthetic, and real-world datasets. The synthetic and semi-synthetic datasets\footnote{The downloadable versions of both synthetic datasets and their simulation can be accessed on our GitHub\footnotemark[1].}, which include true treated and control outcomes, are used to evaluate the treatment effect estimation. In real-world settings, both outcomes are not available. The statistics for the synthetic datasets are described in Table \ref{tb2}.

\subsubsection{Synthetic dataset}
We simulate a synthetic dataset, following existing works \cite{lee2020robust, argaw2022identifying}. The dataset is inspired by the initial clinical trial results of remdesivir to COVID-19 \cite{wang2020remdesivir}. It consists of 10 covariates, which are randomly generated from Normal distribution with different parameter values. The outcomes are simulated using the 'Response Surface B' \cite{hill2011bayesian}. In total, we generate 1,000 samples, with an even split of 500 case and 500 control samples.

\subsubsection{Semi-synthetic dataset}
For the semi-synthetic dataset, we utilize the Infant Health and Development Program (IHDP) dataset, which was originally collected from a randomized experiment aimed at evaluating the impact of early intervention on reducing developmental and health problems among low birth weight, premature infants. The dataset consists of 608 control patients and 139 treated patients, totaling 747 individuals, with 25 covariates. The outcomes are simulated based on the real covariates using the 'Response Surface B' \cite{hill2011bayesian}.

\subsection{Experimental setup}
\subsubsection{Baseline models}
To assess the prediction performance of SubgroupTE, we conduct comparison experiments with two machine learning-based models, support vector machine (SVR) and random forest (RF), as well as state-of-the-art neural network-based models. The following provides a brief description of these models:

\begin{itemize}
\item \textbf{DragonNet~\cite{shi2019adapting}} consists of a shared feature network and three auxiliary networks that predict propensity score, and treated and control outcomes, respectively.
\item\textbf{TARNet~\cite{shalit2017estimating}} consists of a shared feature network and two auxiliary networks that predict treated and control outcomes, respectively.
\item \textbf{VCNet~\cite{nie2021vcnet}} uses separate prediction heads for treatment mapping to preserve and utilize treatment information.
\item \textbf{TransTEE~\cite{zhang2022exploring}} leverages the transformer to model the interaction between the input covariate and treatment.
\end{itemize}

We also conduct comparative experiments with two representative existing subgrouping models to further evaluate the effectiveness of SubgroupTE in subgroup identification. \textbf{R2P}\cite{lee2020robust} is a tree-based subgrouping method, which utilizes a pre-trained treatment effect estimation model to estimate potential outcomes for subgroup identification. \textbf{HEMM} \cite{nagpal2020interpretable} utilizes Gaussian mixture distributions to learn subgroup probabilities. HEMM also employs a pre-trained estimation model to learn the parameters of the distributions. R2P and HEMM use the CMGP\cite{alaa2017bayesian} model and a neural network-based model as treatment effect estimation models, respectively.

\begin{table}[t]
\centering
\caption{Statistics on the synthetic and semi-synthetic datasets}\label{tb2}
\begin{tabular}{ll|cc}\hline
Dataset &         & Synthetic & Semi-synthetic \\ \hline
Number of samples                                                                     & Total   & 1000      & 747            \\
& Case    & 500       & 139            \\
& Control & 500       & 608            \\ \hline
\multirow{2}{*}{\begin{tabular}[c]{@{}l@{}}Avg. of \\ potential outcome\end{tabular}} & Total   & 3.93      & 3.18           \\
& Case    & 2.34      & 6.47           \\
& Control & 5.48      & 2.42           \\ \hline
Number of features                                                                    &         & 10        & 25             \\ \hline
\end{tabular}
\end{table}

\subsubsection{Implementation details}
\begin{figure}[h]
\centering
\includegraphics[width=0.5\linewidth]{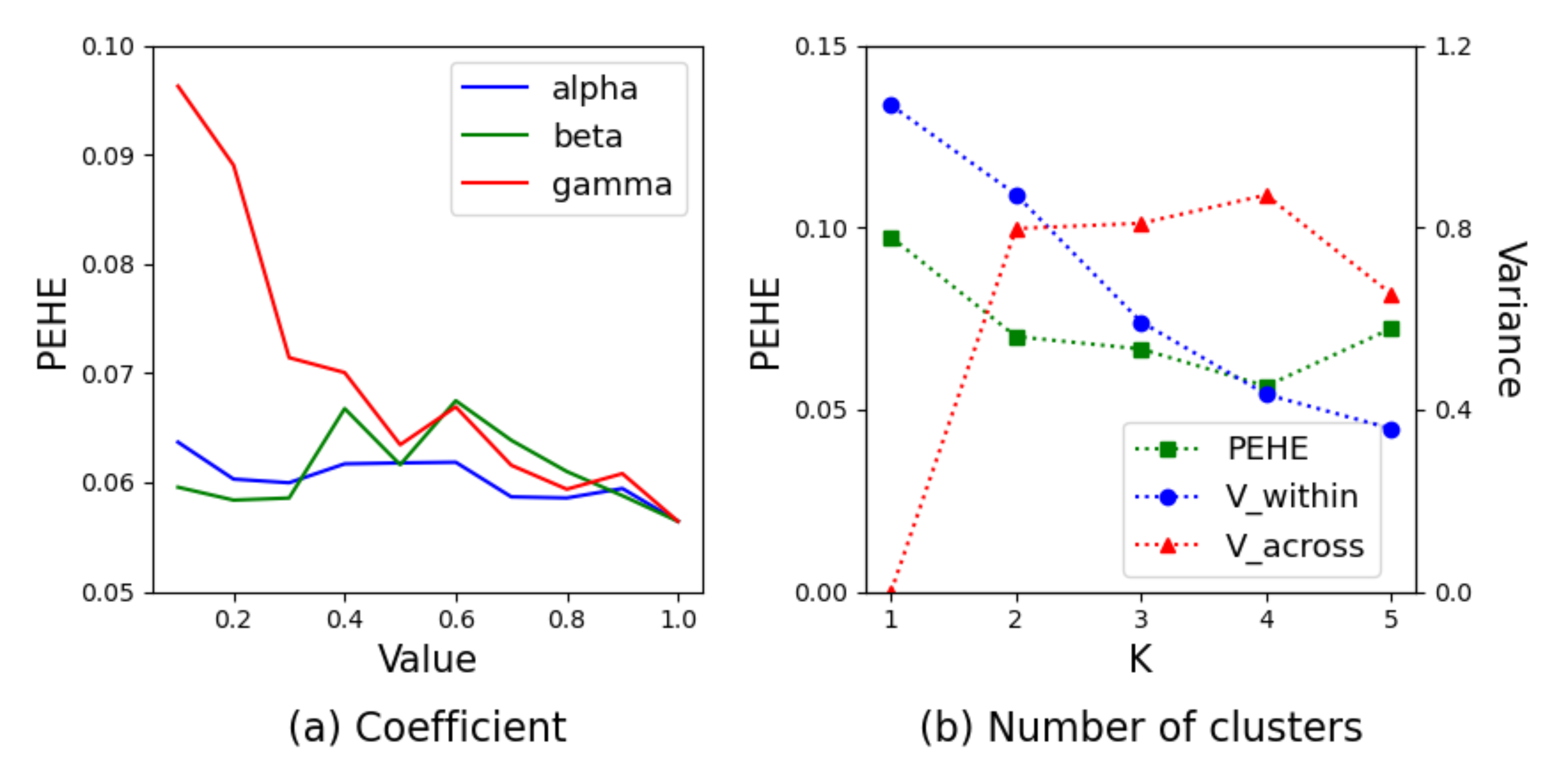}
\caption{Sensitivity analysis conducted for (a) Coefficient and (b) Number of subgroups on the semi-synthetic dataset. For (a), the performance of each coefficient is evaluated while fixing the remaining two coefficients at 1.}\label{fig2}
\end{figure}

All neural network-based models are implemented using PyTorch. We use the SGD optimizer, with a learning rate set to 0.001 and a batch size of 64. The hyper-parameters for the baseline models follow the implementations provided by the respective authors. For our proposed model, we set the hidden nodes from the set \{50, 100, 200, 300\}, the number of subgroups from the range of [1, 10], and coefficients, which are $\alpha$, $\beta$, and $\gamma$, from the range of [0, 1]. The sensitivity analysis for the coefficients and the number of subgroups is shown in Fig. \ref{fig2}. 

We perform experiments with a train : dev : test data split ratio of 6 : 2 : 2. The optimal hyper-parameters are found based on the best performance on the validation data. The evaluation metrics are then reported on the test set, providing an unbiased assessment of the model's performance. 
We provide a description of optimal hyper-parameters, as well as the computing infrastructure, along with the time/space complexity analysis of the model on our GitHub\footnotemark[1].

%We provide a description of the computing infrastructure used and all the optimal hyper-parameters on our GitHub\footnotemark[1].

\subsubsection{Evaluation metric} We employ the precision in estimating heterogeneous effects (PEHE) metric to measure the treatment effect at the individual level, as expressed in Eq. (\ref{pehe}). Additionally, we employ the absolute error in average treatment effect ($\epsilon$ATE) to assess the overall treatment effect at the population level, as defined in Eq. (\ref{ate}). 

% \liucomment{$(f_{y_1}(\mathbf{x}_i)-f_{y_0}(\mathbf{x}_i)$?}
\begin{align}
\label{pehe}
\textrm{PEHE} &= \frac{1}{N}\sum_{i=1}^N (f_{y_1}(\mathbf{x}_i)-f_{y_0}(\mathbf{x}_i)-\mathbb{E}[y_1-y_0|\mathbf{x}_i])^2 \\
\label{ate}
\textrm{$\epsilon$ATE} &= \left |\mathbb{E}[f_{y_1}(\mathbf{x})-f_{y_0}(\mathbf{x})]-\mathbb{E}[y_1-y_0]  \right |
\end{align}

To evaluate the performance for subgroup identification, we analyze the variance of treatment effects within and across subgroups. The variance across the subgroups, $V_{across}$, evaluates the variance of the average treatment effect in each subgroup. On the other hand, the variance within subgroups, $V_{within}$, measures the mean of the variance of the treatment effects in each subgroup. These metrics are defined as Eq. (\ref{across}) and Eq. (\ref{within}), respectively. Here, $TE_k$ indicates a set of the treatment effects in subgroup $k$.

\begin{align}
\label{across}
V_{across} &= Var\left (\{Mean(TE_k)\}_{k=1}^K\right ) \\
\label{within}
V_{within} &= \frac{1}{K}\sum_{k=1}^KVar(TE_k)
\end{align}

\begin{table*}[]
\centering
\caption{Comparison of prediction performance on the synthetic and semi-synthetic datasets. The average score and standard deviation under 30 trials are reported.}\label{tb3}
\begin{tabular}{ll|cc|cc}\hline
\multicolumn{2}{c|}{Dataset}     & \multicolumn{2}{c|}{Synthetic}                                                            & \multicolumn{2}{c}{Semi-synthetic}                                                        \\
\multicolumn{2}{c|}{Model}       & PEHE                                        & $\epsilon$ATE                                         & PEHE                                        & $\epsilon$ATE                                         \\ \hline
ML & RF         & 0.086 $\pm$ 0.000                           & 0.039 $\pm$ 0.000                           & 0.179 $\pm$ 0.000                           & 0.095 $\pm$ 0.020                           \\
                    & SVR        & 0.103 $\pm$ 0.000                           & 0.029 $\pm$ 0.000                           & 0.198 $\pm$ 0.000                           & 0.112 $\pm$ 0.023                           \\ \hline
DL                  & DragonNet  & 0.081 $\pm$ 0.013                           & 0.016 $\pm$ 0.013                           & 0.105 $\pm$ 0.037                           & 0.040 $\pm$ 0.010                           \\
                    & TARNet     & 0.068 $\pm$ 0.010                           & 0.023 $\pm$ 0.003                           & 0.092 $\pm$ 0.019                           & 0.039 $\pm$ 0.010                           \\
                    & VCNet      & 0.034 $\pm$ 0.005                           & 0.018 $\pm$ 0.010                           & 0.080 $\pm$ 0.031                           & 0.065 $\pm$ 0.049                           \\
                    & TransTEE   & 0.045 $\pm$ 0.046                           & 0.026 $\pm$ 0.055                           & 0.099 $\pm$ 0.071                           & 0.153 $\pm$ 0.046                           \\
                    & SubgroupTE & \textbf{0.024 $\pm$ 0.002} & \textbf{0.014 $\pm$ 0.009} & \textbf{0.056 $\pm$ 0.018} & \textbf{0.039 $\pm$ 0.037} \\ \hline
\end{tabular}
\end{table*}

\subsection{Results on synthetic data}

\subsubsection{\textbf{Treatment effect estimation}}

Table~\ref{tb3} shows the prediction performance on the synthetic and semi-synthetic datasets. Our proposed SubgroupTE achieves the best performance in both PEHE $\epsilon$ATE compared to other baselines. Notably, it achieves the PEHE of 0.024 and 0.056 on the synthetic and semi-synthetic datasets, respectively, which shows a reduction of 29.4 \% and 30.0 \% compared to the second-best model. The results demonstrate the effectiveness of SubgroupTE, which integrates subgroup information in the estimation process, thereby enhancing treatment effect estimation.

\subsubsection{\textbf{Subgroup identification}} 
We evaluate whether the model identifies subgroups that maximize heterogeneity between different subgroups while ensuring homogeneity within each subgroup.

\begin{figure}[]
\centering
\includegraphics[width=0.5\linewidth]{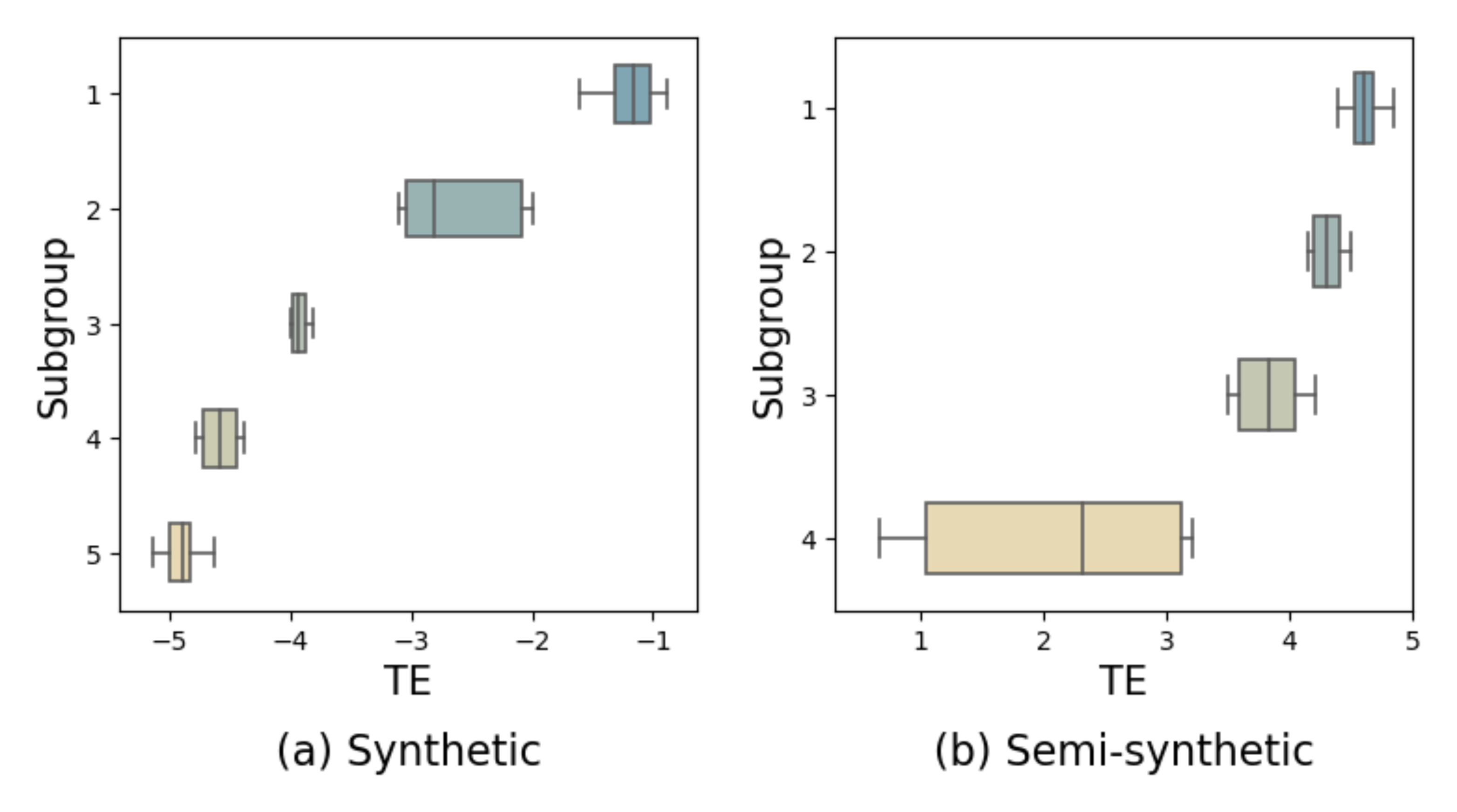}
\caption{The boxplots of the treatment effect distribution for the identified subgroups on the (a) synthetic and (b) semi-synthetic datasets. The box spans from the first quartile to the third quartile of the data, with a line indicating the median. The whiskers extend from the box to encompass the 5th to 95th percentiles.}\label{fig3}
\end{figure}

Table \ref{tb4} presents the results of the subgrouping performance on the semi-synthetic dataset. SubgroupTE achieves the best performance
compared to all baselines. Both baseline models have lower performance in PEHE, indicating that the pre-trained estimation models are not performing well. As a result, the subgroup performance metrics are shown to be inferior to SubgroupTE. This highlights the limitations of existing subgrouping methods that rely on pre-trained models. In contrast, SubgroupTE demonstrates improved performance in both subgrouping and treatment effect estimation by iteratively training these tasks. Additionally, Fig. \ref{fig3}(b) shows the sensitivity analysis on the number of subgroups. The results demonstrate the significance of identifying subgroups with similar treatment responses and integrating them into the estimation process for treatment effect estimation. When the number of subgroups is 1, treating the entire population as a single subgroup, PEHE is approximately 0.1, similar to other baseline methods. However, as the number of subgroups increases, the PEHE decreases, with the best performance observed when there are 4 subgroups. This further evaluates the effectiveness of our proposed method in effectively identifying diverse subgroups and accurately estimating subgroup-specific treatment effects.

\begin{table}[!t]
\centering
\caption{Comparison of subgrouping performance on the semi-synthetic dataset.}\label{tb4}
\begin{tabular}{l|ccc}\hline
\multicolumn{1}{c|}{Model} & $V_{within}\downarrow$      & $V_{across}\uparrow$      & PEHE $\downarrow$           \\ \hline
R2P                        & 0.500$\pm$0.15 & 0.643$\pm$0.13 & 0.154$\pm$0.05 \\
HEMM                       & 0.570$\pm$0.11 & 0.591$\pm$0.15 & 0.172$\pm$0.00  \\
SubgroupTE                 & \textbf{0.393$\pm$0.02} & \textbf{0.901$\pm$0.01} & \textbf{0.056$\pm$0.02}     \\ \hline
\end{tabular}
\end{table}

To evaluate whether the model identifies heterogeneous subgroups, we visualize the distributions of the true treatment effects from all subgroups. As shown in Figure~\ref{fig3}, SubgroupTE reliably distinguishes heterogeneous subgroups, showing substantial variations in the average treatment effects and clearly separated distributions across different subgroups. The results provide evidence that supports the usefulness of SubgroupTE for subgroup identification, demonstrating its ability to capture and account for the inherent heterogeneity in treatment response within the population.

\begin{figure}[h]
\centering
\includegraphics[width=0.4\linewidth]{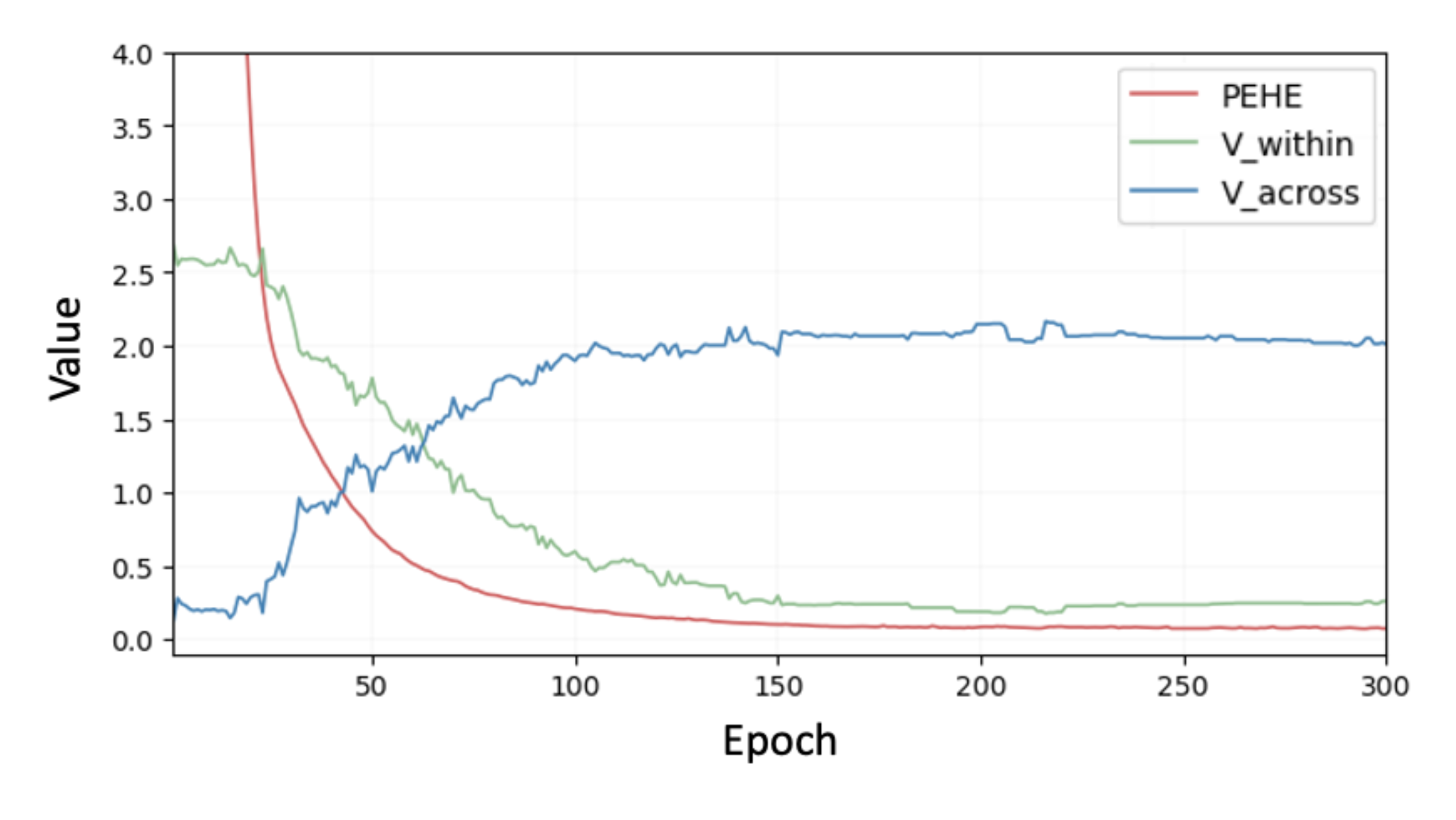}
\caption{Illustration of the trends in PEHE and variance within and across subgroups during
the training phase on the validation set of the synthetic dataset.}\label{fig4}
\end{figure}

SubgroupTE iteratively updates cluster centroids and network parameters until they are converged to continuously enhance subgroup identification. Fig. \ref{fig4} illustrates the trends in PEHE and variance within and across subgroups. As depicted, the variance within subgroups decreases over epochs, while the variance across subgroups increases. These results provide empirical evidence of the continuous improvement in subgroup identification through iterative optimization. Furthermore, it is notable that PEHE decreases over epochs. This demonstrates the mutually reinforcing nature of subgroup identification and treatment effect estimation, indicating that both aspects are effectively learned and improved throughout the training process.

\subsection{Ablation study}
We conduct an ablation study to investigate the effectiveness of the EM-based iterative training process. We compare two variants. \textbf{SubgroupTE-O} follows; 1) The model is pre-trained without subgrouping, 2) The cluster centroids are updated and fixed, and 3) The model is re-trained using the subgroup probability derived from the fixed centroids. \textbf{SubgroupTE-P} follows; 1) The model is initially pre-trained without subgrouping, and 2) The model parameters and the centroids are iteratively updated. As shown in Table \ref{tb5}, SubgroupTE achieves the highest performance. These findings provide evidence for the effectiveness of our training process, where subgrouping and estimation are learned in a mutually reinforcing manner. Additionally, the results clearly demonstrate that iterative optimization for subgrouping and treatment estimation enhances estimation accuracy and subgroup identification, overcoming the limitations of the one-time estimation in the existing subgrouping models.

\begin{table}[!t]
\centering
\caption{Ablation study for SubgroupTE on the semi-synthetic dataset.}\label{tb5}
\begin{tabular}{l|ccc} \hline
\multicolumn{1}{c|}{Model} & $V_{within}\downarrow$      & $V_{across}\uparrow$      & PEHE $\downarrow$       \\ \hline
SubgroupTE-O               & 0.452$\pm$0.00 & 0.854$\pm$0.00 & 0.063$\pm$0.00 \\
SubgroupTE-P               & 0.453$\pm$0.01 & 0.878$\pm$0.01 & 0.060$\pm$0.01 \\
SubgroupTE   & \textbf{0.393$\pm$0.02} & \textbf{0.901$\pm$0.01} & \textbf{0.056$\pm$0.02} \\ \hline
\end{tabular}
\end{table}

\subsection{Real-world study}
\subsubsection{Problem statement and dataset} Opioid use disorder (OUD) imposes a substantial healthcare and economic burden. Despite various Food and Drug Administration (FDA)-approved drugs for OUD treatment, including Methadone, Buprenorphine, and Naltrexone, they are either restricted in usage or less effective. Naltrexone is accessible through any licensed medical practitioner, whereas Methadone and Buprenorphine require regulatory compliance\cite{sharma2017update}; Methadone is exclusively accessible through regulated opioid treatment programs (OTPs), and Buprenorphine is prescribable by physicians who have completed specific training or possess addiction board certification and have obtained a federal waiver. Additionally, there exists substantial empirical evidence that supports the effectiveness of medications for OUD (MOUD), particularly Methadone and Buprenorphine\cite{heidbreder2023history}.

\begin{table}[b]
\caption{Statistics on the opioid dataset}\label{tb6}
\centering
\begin{tabular}{l|ccc}
\hline
                & $y=0$ & $y=1$ & Total \\ \hline
Case ($t=1$)    & 403   & 353   & 756   \\
Control ($t=0$) & 1,430 & 707   & 2,137 \\
Total           & 1,833 & 1,060 & 2,893 \\ \hline
\end{tabular}

\end{table}

In light of these regulatory and effectiveness considerations, this study aims to compare Naltrexone, a newly approved treatment, with Methadone and Buprenorphine, well-established drugs, to evaluate Naltrexone's relative efficacy and appropriateness for the treatment of OUD. To conduct this evaluation, the study includes approximately 600,000 patients with OUD from the MarketScan Commercial Claims and Encounters (CCAE) dataset\cite{market}, spanning the years 2012 to 2017. The dataset provides comprehensive medical histories for patients, including prescriptions, diagnoses, procedures, and demographic characteristics. We identify OUD patients based on opioid-related emergency department (ED) visits. 

\subsubsection{Study design} 
\begin{figure}[!t]
\centering
\includegraphics[width=0.7\linewidth]{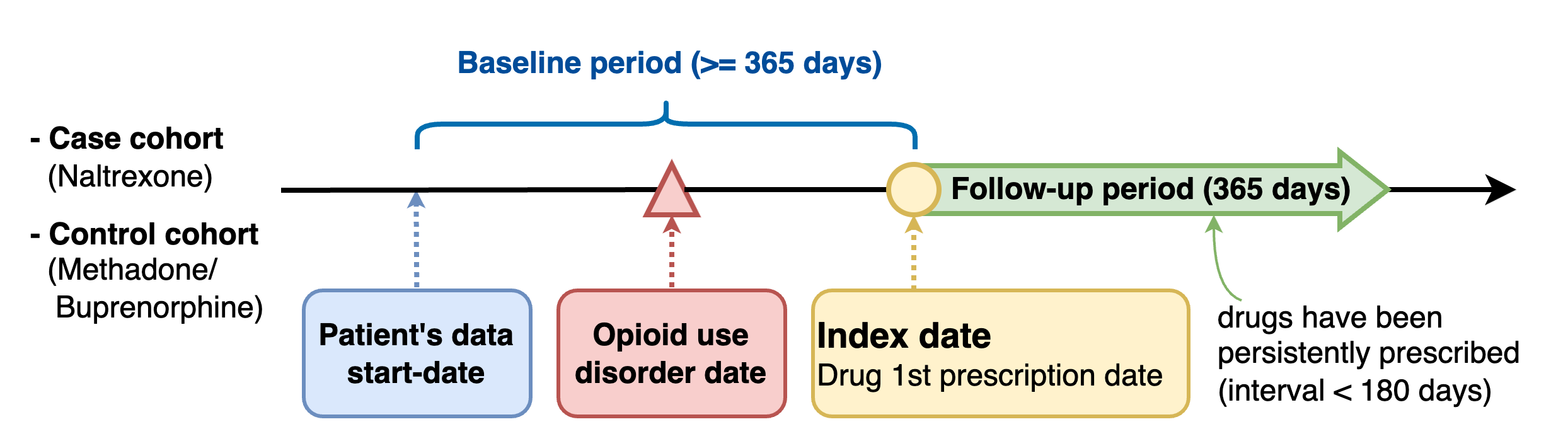}

\caption{Illustration of the cohort selection criteria. The index date indicates the first prescription date of the drug. The baseline and follow-up periods encompass all dates before and after the index date, respectively.}\label{fig5}
\end{figure}

The primary objective of the real-world study is to assess the effect of Naltrexone, compared to Methadone and Buprenorphine, and find subgroups of patients at varying levels of risk for OUD-related adverse events. We collaborated with domain experts 
to define the adverse outcomes associated with OUD, drawing insights from various studies \cite{urman2021burden, sun2022evaluation, zhang2022examining, powell2023variation}. These events include opioid-related adverse drug events (ORADEs), opioid overdose, and hospitalization. Hospitalization refers to whether a patient is admitted to the hospital after using the drug, and ORADEs and opioid overdose are determined by the presence of corresponding diagnosis codes. Patients who experience any of these adverse events are categorized as "positive," while those who do not are labeled as "negative." To assess Naltrexone's relative effect compared to Methadone and Buprenorphine, we define case and control cohorts. A case cohort consists of patients prescribed Naltrexone, whereas a control cohort comprises patients prescribed Methadone or Buprenorphine. Detailed selection criteria and statistics are provided in Table \ref{tb6}, and Fig. \ref{fig5} illustrates the specific criteria.

We construct pre-treatment covariates using diagnosis and medication codes. The diagnosis codes, which are International Classification of Diseases (ICD) codes, are mapped into Clinical Classifications Software (CCS), resulting in 286 unique codes. Additionally, national drug codes (NDCs) of medications are aligned with observational medical outcomes partnership (OMOP) ingredient concept IDs, encompassing 1,353 different drugs. To select the optimal number of subgroups ($K$), we consider both $V_{across}$ and $V_{within}$.

\subsubsection{Results} In Fig. \ref{fig3} (b), we present a visualization of the treatment effect distribution for the identified subgroups. Among the identified subgroups, the first subgroup exhibits a positive average treatment effect, suggesting a diminished effect for Naltrexone within this subgroup. The second subgroup shows an average treatment effect that is close to zero, implying that Naltrexone has minimal impact on the outcomes for these patients. On the other hand, the third subgroup stands out as a notably negative average treatment effect, suggesting that Naltrexone has an enhanced effect compared to Methadone or Buprenorphine. Consequently, Naltrexone might be the preferable treatment option for the third subgroup. Based on these observations, Naltrexone would be most recommended as a treatment option for the third subgroup, where it appears to have the most substantial enhanced effect.

\begin{figure}[!th]
\centering
\includegraphics[width=0.3\linewidth]{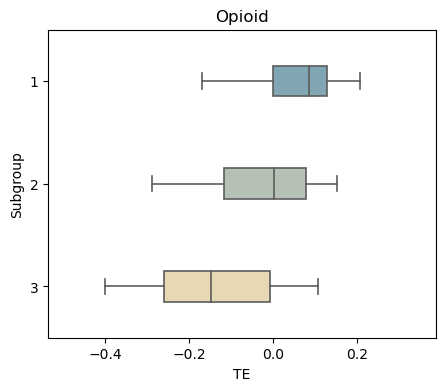}
\caption{The boxplots of the treatment effect distribution for the identified subgroups on the opioid dataset. }\label{fig6}
\end{figure}

\begin{figure}[!th]
\centering
\includegraphics[width=0.6\linewidth]{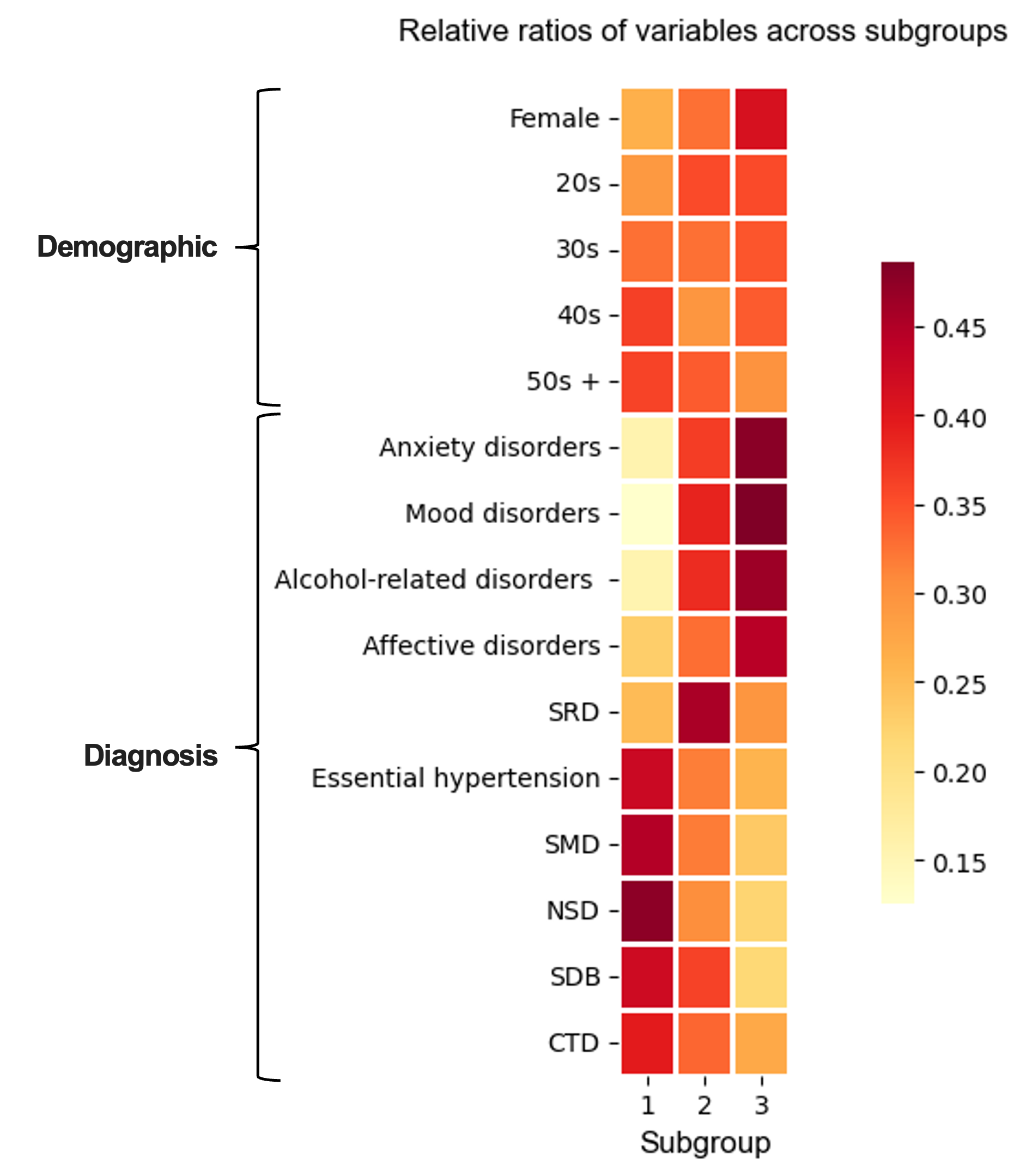}
\caption{The heatmap of the relative ratios of variables for demographics and diagnosis codes among the three subgroups. These relative ratios are calculated using the formula $\pi_{k,i}/\sum_{k=1}^{K}\pi_{k,i}$, where $\pi_{k,i}$ represents the ratio of the $i$-th variable within the $k$-th subgroup. SMD: Substance-related mental disorders; NSD: Other nervous system disorders; SDB: Spondylosis, intervertebral disc disorders, or other back problems; CTB: Other connective tissue disease.
%\liucomment{add descriptions of the scaling/normalization here}
}\label{fig7}
\end{figure}

To perform a more comprehensive comparison of the identified subgroups, we analyze the variables related to diagnosis codes and demographics for all subgroups. In Fig. \ref{fig7}, we present a heatmap displaying the relative ratios of the variables for the three subgroups. The ratios are calculated by determining the ratio of each variable within each subgroup and then scaling these ratios across all subgroups. We analyze diagnosis codes recorded only during the baseline period. From the top 15 most frequent codes, we report the 10 codes that exhibit the most differentiation across subgroups. Our analysis reveals intriguing patterns: as the treatment effect improves, the ratios of diagnosis codes typically show a gradual increase or decrease Furthermore, the third subgroup, which experiences the most significant treatment enhancement, is characterized by a higher proportion of females and younger patients. The findings provide strong evidence that the identified subgroups are indeed clinically distinct. Moreover, they underscore the efficacy of our proposed model in accurately identifying subgroups.

The results indicate the strength of SubgroupTE in identifying subgroups with heterogeneous treatment effects. This demonstrates the advantages of SubgroupTE in developing personalized treatment strategies by not only estimating treatment effects but also recommending a specific treatment for each subgroup. Furthermore, Fig. \ref{fig7} shows that SubgroupTE effectively understands the characteristics of each subgroup based on patients' medical history and helps to identify the variables contributing to the improvement of treatment effects.

\section{Conclusion}
This study addresses the critical problems related to treatment effect estimation. We propose a novel framework that incorporates subgrouping and treatment effect estimation to account for the heterogeneity of responses within the population. Through subgroup identification, our model learns subgroup-specific causal effects, thereby advancing treatment effect estimation. We demonstrate the effectiveness of our approach through extensive experiments and analysis. Our model outperforms the state-of-the-art baselines and shows superior performance in both treatment effects estimation and subgroup identification. Furthermore, experiments on the real-world dataset demonstrate the potential of our framework in enhancing treatment recommendation and optimization in clinical practice.

%Bibliography
\bibliographystyle{unsrt}  
\bibliography{references}

\end{document}